\documentclass[final,5p,times,twocolumn]{elsarticle}

\usepackage{graphicx}
\usepackage{amssymb}
\usepackage{scrextend}
\usepackage{xcolor}
\usepackage{amsmath}
\usepackage{enumitem}
\usepackage{pgfplots}
\usepackage{pdfpages}
\usepackage{url}
\usepackage{multirow}
\usepackage{tikz}
\usepackage{dblfloatfix}
\usepackage{array}
\usepackage{soul}
\newcolumntype{x}[1]{>{\centering\let\newline\\\arraybackslash\hspace{0pt}}p{#1}}
\usepackage{hyperref}

\definecolor{bblue}{HTML}{4F81BD}
\definecolor{rred}{HTML}{C0504D}
\definecolor{ggreen}{HTML}{9BBB59}
\definecolor{ppurple}{HTML}{9F4C7C}
\definecolor{deepjunglegreen}{rgb}{0.0, 0.29, 0.29}
\usepackage{color}

\journal{Computational Materials Science}

\begin{document}

\begin{frontmatter}


\title{MatScIE: An automated tool for the generation of databases of methods and parameters used in the computational materials science literature}


\author[label1]{Souradip Guha\corref{cor1}}
\ead{sourya2793@gmail.com}
\author[label1]{Ankan Mullick\corref{cor1}}
\ead{ankanm@kgpian.iitkgp.ac.in}
\author[label1]{Jatin Agrawal}
\ead{Jatin86400@iitkgp.ac.in}
\author[label2]{Swetarekha Ram}
\ead{swetarekha.ram@ikst.res.in}
\author[label1]{Samir Ghui}
\ead{samirghui@iitkgp.ac.in}
\author[label2]{Seung-Cheol Lee\corref{cor2}}
\ead{seungcheol.lee@ikst.res.in}
\author[label2]{Satadeep Bhattacharjee\corref{cor2}}
\ead{s.bhattacharjee@ikst.res.in}
\author[label1]{Pawan Goyal\corref{cor2}}
\ead{pawang@cse.iitkgp.ac.in}
\cortext[cor1]{Authors contributed equally to this work.}
\cortext[cor2]{Corresponding Author}

\address[label1]{Indian Institute of Technology, Kharagpur}
\address[label2]{Indo-Korea Science and Technology}


\vspace{-5mm}
\begin{abstract}

The number of published articles in the field of materials science is growing rapidly every year.
This comparatively unstructured data source, which contains a large amount of information, has a restriction on its re-usability, as the information needed to carry out further calculations using the data in it must be extracted manually. It is very important to obtain valid and contextually correct information from the online (offline) data, as it can be useful not only to generate inputs for further calculations, but also to incorporate them into a querying framework. Retaining this context as a priority, we have developed an automated tool, MatScIE (Material Science Information Extractor) that can extract relevant information from material science literature and make a structured database that is much easier to use for material simulations. Specifically, we extract the material details, methods, code, parameters, and structure from the various research articles. Finally, we created a web application where users can upload  published articles and view/download the information obtained from this tool and can create their own databases for their personal uses.
\end{abstract}


\end{frontmatter}

\section{Introduction}
\label{S:1}

Currently, the majority of material science findings and information are stored in an unstructured format across {numerous} published articles. 
A typical article contains information on the material(s) studied, the method used, the computational software(s) used in the study, the simulation parameters and, finally, the outcome of the study.
If we consider the scenario where we want to query the methods and parameters discussed in published material science papers, there is no easy and robust method to effectively filter and refine this information automatically. Manually going through the papers and finding out the methods used is an inefficient and tedious task. A potential solution can be to build a system that can automatically extract mentions of a method from any published article.

To solve these problems, we introduce Material Science Information Extractor (MatScIE), which is capable of extracting information about the material, code, parameter, method and structure from the published research article, along with providing a summary of the main research findings. 
Significant developments have been observed in the field of information extraction using machine learning and deep learning techniques. A very specific and widely used task is Named Entity Recognition (NER) that classifies (extracts) named entities in the text as per pre-decided classes or categories. It accepts as an input a sequence of tokens and identifies the spans in the input sequence that belong to one of the pre-decided categories. In our use-case, we attempt to extract the spans of text in a material science research article belonging to one of the five categories: \textit{material, code, parameter, method, structure}. This enables us to adapt the NER framework for our task. 

Popular NLP methods for the NER task are based on recent advances in deep-learning \cite{huang2015bidirectional}. An important requirement in training a deep learning model is the availability of appropriate (and large) annotated data. We created a modest annotated dataset using 214 material science articles by labeling the text spans into the five categories. 

In the first part, we trained standard sequential models on the supervised data to predict labels. We utilized word embeddings pre-trained in the material science domain.
In the second part, we injected noise in the training dataset to increase the robustness of the model. Since injecting noise in the textual dataset is critical, we developed a novel procedure using a {\sl Relabeling and Mimicking model}. The Relabeling model has a high recall, we use this model to inject noise in the dataset. We varied the amount of noise injected, and obtained results pertaining to different amount of injected noise. In order to derive a short summary level information from the published article, we trained a sentence classification model on an annotated dataset of 90 articles. Additionally we developed a web application that generates summary level information and token spans corresponding to each category.

The remainder of the paper is structured as follows. In Section \ref{S:2}, we describe past research on information extraction on scientific articles. We describe our annotated dataset in Section \ref{S:3}. Then we describe our proposed approach and evaluation metrics in Sections \ref{S:4} and \ref{S:5}, respectively. We show the experimental results and sample outputs in Sections \ref{S:6} and \ref{S:7}, respectively. Some more analysis on the output are presented in Section \ref{S:8}. We present the online interface in Section \ref{S:9}. Section \ref{S:10} concludes the paper and gives directions for the future work.

\vspace{-3mm}
\section{Related Work}
\label{S:2}
\vspace{-2mm}

Information extraction from scientific articles has been extensively explored recently. With the help of information extraction methods we intend to extract potential information out of the large body of scientific articles. Extracted information can provide an overview of the key-insights from scientific articles. Some of the major computational approaches in this regards include rule-based approaches, machine learning approaches like Naive Bayes classifier \cite{xu2018bayesian}, support vector machines \cite{alsaleem2011automated} and deep learning approaches. Most of the studies with regards to deep learning are performed on publicly available datasets like bc5cdr \cite{li2016biocreative} and SCIERC \cite{luan2018multi}. One interesting work is this field is done by \citet{luan2018multi}, that extracted entities, relations between entities and co-reference clusters in scientific articles using a multi-task setup. They applied their model on SCIERC dataset, that contains abstracts of 500 scientific articles from different domains. Furthermore they used their predictions to generate a scientific knowledge graph that can further be used to show analysis on scientific literature. Another popular work done in this field is that of \citet{beltagy2019scibert}. This work uses a BERT model pretrained on a corpus consisting of scientific articles to improve performance on downstream scientific NLP tasks. Corpus used for pretraining the model mostly consisted of biomedical documents. Since our work is primarily based on material science domain articles, we generated an annotated dataset consisting of only material science articles and then trained deep learning models on the same.

There has been quite some work to extract Chemical entities. Some of the information related to chemical entities is already available in static databases. These databases map chemical information to relevant document with details like patents, literature etc related to the entered text. Below is a list of popular chemical databases :
\begin{itemize}
   \item  PubChem~\cite{wang2009pubchem}: It is an open chemistry database where researchers can add scientific data that others may use. It is maintained by NCBI~\cite{geer2010ncbi}. It contains molecule names, substance descriptions and link to published documents related to the queried chemical compound.
   \item  ChemSpider~\cite{pence2010chemspider} - It is a free database containing chemical structures with the feature to search by chemical names and chemical structures. It helps to find important data like literature references, physical properties, chemical suppliers etc.
   \item SciFinder~\cite{ridley2009information} - It is used to access information from selected Chemical Abstract Services(CAS) databases. It offers searches of authors name, related topic etc.
\end{itemize}
But all such databases are static and require constant updates from researchers.

One interesting work in this field is the OSCAR4 recogniser \cite{jessop2011oscar4}. It build an n-gram model and then uses it in Bayesian classifier to classify whether a token belongs to ``chemical" or ``non-chemical". The n-gram model is built with the help of a list of chemical tokens, that are obtained from a fixed dictionary and manually annotated documents. It then builds a Maximum Entropy Markov Model~\cite{mccallum2000maximum} by representing each token with a set of features, one of which is the probability that it belongs to a chemical domain with the predictions of the previously built n-gram model. Since the model is prepared from a fixed dictionary, more often than not it fails to capture the materials when represented in complex notations. Additionally, this model extracts only the chemical components from a text whereas we are interested in a much detailed extraction from the text. 

ChemSpot~\cite{rocktaschel2012chemspot} is another chemical extraction tool that uses Conditional Random Field(CRF)~\cite{lafferty2001conditional} to identify IUPAC and IUPAC-like names from a scientific text. This model also fails to fulfill our purpose for similar reasons.

Recently there are research studies being performed on material science domain articles which use deep learning models to classify tokens into material science categories.
One such work is that of Weston et al. \cite{weston2019named}. It uses BiLSTM with CRF to classify each token of the text. Dataset used in this paper contains 800 article abstracts which are annotated among the following labels - \textit{material, symmetry/phase label, sample descriptor, property, application, synthesis method and characterization method}. However, the entity labels used in that paper is different from the labels in which we are interested in since we are dealing with papers that report density functional theory / first principle-based calculations. 

Several researchers have focused on traditional machine learning models to explore textual features to perform several tasks like understanding material science languages, extract information or develop a knowledge graph. One such work was done by Tshitoyan et al.~\cite{tshitoyan2019unsupervised} who developed an unsupervised word embedding model to process text for identifying complex materials science knowledge such as the underlying structure of the periodic table and structure–property relationships in materials\footnote{We have used this embedding technique as a baseline in our experiment.}.  
Another work was done by Buttler et al.~\cite{butler2018machine}. They have explored the evolution of research workflow and how different machine learning algorithms (like Genetic Algorithms, Na\"{i}ve Rules based Approaches\footnote{We have used Na\"{i}ve Rule approach as a baseline} etc.) would be used to design, synthesize different aspects of chemical and material science research areas. Hakimi et al. \cite{hakimi2020time} focused on biomaterial text mining for retrieving relevant documents. Kostoff et al.~\cite{kostoff2005method} described techniques for data retrieval and discussed the desirability of conflating search terms. Another direction of work was to develop an automatic approach. Kim et al.~\cite{kim2017materials} built a platform using variety of machine learning and natural language processing techniques to automatically retrieve
articles and then extract materials synthesis conditions found in the text. Juan-Pablo et al.~\cite{correa2018accelerating} presented how machine learning and natural language processing can help solving long timelines and low success probabilities of material science research and can accelerate the pace of novel materials development. Some researchers were also focused on exploring material science doamin relevant features. One such work was done by Goldsmith et al.~\cite{goldsmith2018machine}. They have shown how machine learning can be useful for aiding heterogeneous catalyst understanding, design and discovery. Dragone et al.~\cite{dragone2017autonomous} proposed a system that can evaluate chemical reactivity and detect for new reactions, rather than a predefined set of targets. 

Few works were also done on material science synthesis procedure.  The semi-supervised machine-learning classification model by Huo et al.~\cite{huo2019semi} used a Dirichlet allocation model to cluster keywords into topics corresponding to specific experimental materials synthesis steps, such as “grinding” and “heating”, “dissolving” and “centrifuging”, etc. Guided by a modest amount of annotation, a random forest classifier can then associate these steps with different categories of materials synthesis, such as solid-state or hydrothermal. Some people explored semantic features of text - Mysore et al.~\cite{mysore2019materials}. Young et al.~\cite{young2018data} experimented on data mining techniques in case of pulsed laser deposition of complex oxides. Mysore et al.~\cite{mysore2017automatically} have built graph based automatic extraction model. Kononova et al.~\cite{kononova2019text} have automatically extracted synthesis entries from scientific publications to gather a dataset of “codified recipes” for solid-state synthesis. Himanen et al.~\cite{himanen2019data} worked on data driven approaches on material science articles.

The real complexity of our task lies in extracting only the entities used in the study rather than those that are just being mentioned. 
For example, consider the paper with title, \textit{``Temperature independent band structure of WTe2 as observed from ARPES"}. This paper may mention many different \textit{materials} but only WTe2 is being studied, and is the only \textit{material} in which we are interested in. We observe that an article usually works on a much smaller subset of the \textit{materials} than the ones actually occurred in the paper, while the remaining are just mentioned as previous work. In a normal NER task, however, we would be interested in extracting all the different \textit{materials} mentioned in the paper. The same goes for all other classes.

\section{Dataset}
\label{S:3}
In this section, we discuss the collection and annotation process of the dataset. We first discuss the dataset creation for the information extraction task, followed by the dataset for sentence classification task.

\subsection{Information Extraction Dataset}
\label{S:3.1}
We started creating our dataset using a set of 70 hand-picked material science articles. To expand this, we crawled material science articles published between 2010 to 2019 from arxiv under \textit{cond-mat.mtrl-sci} category.  
We applied filters to consider only the articles that contain atleast one among the following keywords in the abstract : ``ab initio simulation", ``density functional study", ``density functional theory" and ``first principles".  

Next, we considered only those articles with atleast one \textit{code} listed on \url{https://psi-k.net/software}. Thus, we ended up with a repository of 10,900 material science articles in PDF format. 

\subsection{PDF annotation and parsing}
In addition to the initial 70 documents, we took some articles randomly from the arxiv dataset to create a dataset of 214 articles. These PDF documents are annotated using Google Pdf Editor by experts in material science domain. We selected the following entity types for labelling : a) \textit{material} b) \textit{method} c) \textit{code} d) \textit{parameter} e) \textit{structure}. Selective words or series of words were highlighted and annotated with the corresponding label using this annotation tool. If there are multiple occurrences of the same word or series of words, then only one occurrence is labelled. The rest of the occurrences are automatically annotated when we process the data. 

Since the annotations are in the PDF format, we need to extract the contents along with the annotation to be able to apply NLP approaches. To achieve this we analysed several open source applications as listed below.

\begin{itemize}[itemsep=-0.4em]
  \item pdfanno\footnote{\url{https://github.com/paperai/pdfanno}}
  \item brat\footnote{\url{https://brat.nlplab.org}}
  \item INCEpTION\footnote{\url{https://github.com/inception-project/inception}}
  \item popplerqt4\footnote{\url{https://github.com/frescobaldi/python-poppler-qt4}}
  \item OCR++\footnote{\url{https://github.com/ocrplusplus/ocrplusplus}} \cite{singh2016ocr++}
  \item Science Parse\footnote{\url{https://github.com/allenai/science-parse}}
\end{itemize}

To this end, we finalised two applications, \textit{popplerqt4} and \textit{Science Parse}, that are both usable and appropriate to create the dataset, as described below.

\begin{itemize}
\item\textit{popplerqt4} - 
popplerqt4 is a python library that can be used to extract annotations (highlights, comments) from a PDF file, and formats them as markdown text. The output of the script contains each highlighted text and corresponding comments. However, it does not provide the sentence containing the annotation.

\item\textit{Science Parser} - 
Science Parser parses scientific papers (in PDF form) and returns in a structured form. It extracts Title, Authors, Abstract, Sections, Bibliography, Mentions, etc. in a json format. It does not have the ability to extract highlighted portions and comments. 

\end{itemize}

\begin{figure*}
    \centering
 \includegraphics[scale=0.5]{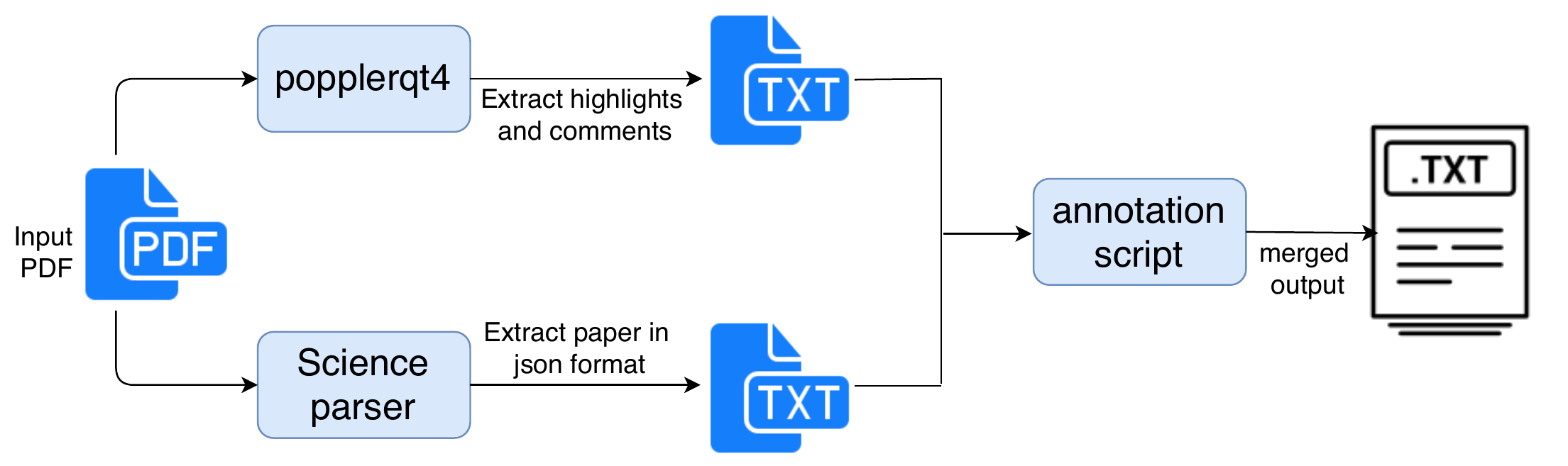}
	\vspace{-2mm}
	\caption{Converting PDF documents to text format while preserving annotations}
    \label{fig:dataset_creation1}
\end{figure*}

We combine the output of \textit{Science Parser} and \textit{popplerqt4} to generate a well formed output containing the contents of scientific article along with annotations. This is done by the \textit{annotation script} that first lists out the annotations of the PDF article using \textit{popplerqt4} and then uses regular expressions to match the annotations with text obtained from \textit{Science Parser} application as shown in Figure  \ref{fig:dataset_creation1}.

\subsection{Text Preprocessing}
The above steps provide the annotated PDF document in raw text format. However, to make the data ready for the extraction task, the first step is to split the raw text into sentences. An adhoc splitting of sentences with respect to period (.) might not be the best choice since such token might occur multiple times within the same sentence. Hence we used \textit{Spacy}, a sentence tokenizer tool, that uses machine learning algorithm to learn the end-of-lines (eol).

Each sentence consists of tokens. As mentioned earlier, a chemical formulae can be represented in multiple forms. To perform normalization of chemical formulae, we replaced all numbers with an uniform digit(0). We extracted the tokens from a sentence by spitting the token based on whitespace. 

Generally the output obtained from \textit{Science Parser} contain spurious characters that are either misrepresented or are added erroneously. We cleaned the text by either removing them or by replacing those with appropriate letters.

\subsection{Selecting sentences}

After tokenization of the text sentences, we have two choices : a) consider only the sentences that contains any token that is labelled b) consider all the sentences. We prepare dataset based on both the choices. Later, we will see that the dataset prepared with the second choice provides better results. Since we are interested in finding spans of sentences belonging to a specific category, there is a need to convert the data to some tagging format. We used inside-outside-beginning (IOB) format which is a very common tagging format used in named entiry recognition (NER) tasks. After all the above mentioned operations, we ended up with 49,610 sentences in total where the tokens are annotated among 5 classes: (\textit{material, method, code, parameter, structure}).  
Any token that is not associated to these classes is treated as if it belongs to a special class ``O".

\subsection{Label Distribution and division of the dataset}

\begin{figure}
    \centering
     \includegraphics[width=\linewidth]{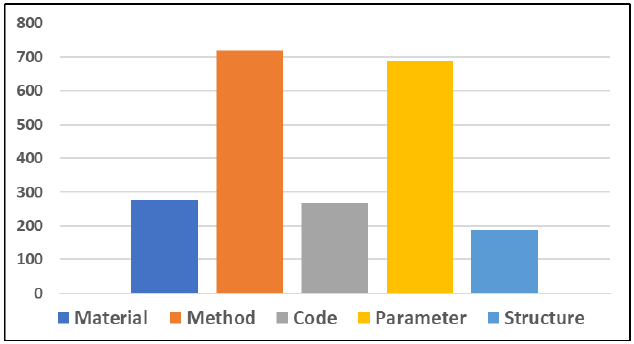}
    \caption{Label distribution of the five different classes.}
    \vspace{-3mm}
    \label{fig:labelDistribution}
\end{figure}

Figure \ref{fig:labelDistribution} shows the number of spans corresponding to each class in the annotated data. Multiple occurrences of a span in the same document is counted once. From the figure it is clear that the labels follow a skew distribution with maximum number of spans observed in \textit{method} \& \textit{parameter} classes and minimum number of spans are observed for \textit{structure} class.

We have performed 5-fold cross validation test and we report average and standard deviation of the precision, recall, F1-score of the model across these 5 folds. Note that we did the division using documents and not sentences to reflect the real world setting, where we would be provided with a new document at test time. Following standard practices, for each fold, the model parameters will be learnt on the training set, the hyper-parameters will be tuned on the validation set, while not touching the test set. The final tuned model will be used on the test set to get the final numbers.

\subsection{Sentence classification Dataset}
\label{sentence_class}
Since the previous annotations would help in extracting the material, methods, code, parameter and structure from the scientific document, a decent summary level information of the article can be derived by including sentences that contain some information related to results. We, therefore, chose abstracts of 90 articles from the 214 articles above and annotated those into \textit{positive} or \textit{negative} depending on whether the corresponding sentence corresponds to a result\footnote{Data available at https://github.com/TeamMatSciE/MatSciE}.

\section{Proposed Approach}
\label{S:4}

Our proposed approach consists of an information extraction and a sentence classification model, which we describe in this section. For information extraction, we utilize a sequence labeling approach, commonly used to identify the (named) entities of interest. Given a sequence of tokens (words), this model makes a judgement for each token in the sequence, as to whether it is part of any of the underlying entities of interest. 

For example, the model should output that a token ``Na2SO4" belongs to the class \textit{material} and the sequence ``Vienna Ab Simulo Package" belongs to the class \textit{code}.

\begin{figure}[!tb]
    \centering
	\includegraphics[width=\linewidth]{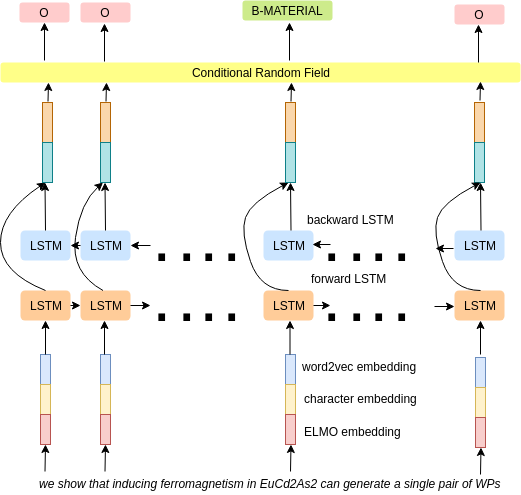}
	\caption{Overview of BiLSTM-CRF model. Each token is represented by the concatenation of word2vec, character and ELMO embedding. In each time step, the model learns both left-to-right and right-to-left context. Finally the output vector is passed to CRF layer to obain the final predicted label.}
	\label{fig:model}
\end{figure}

In the NLP literature, there are primarily three types of information that are utilized for the sequence labeling task: a) word representation, b) the context in which the word has occurred in the sentence, and c) shape of the token. 

Word representation can be obtained by training a word-embedding model. We used the \textit{word2vec} model with \textit{skip-gram} architecture to represent the word in vector as discussed by \citet{mikolov2013efficient}. Skip-gram architecture utilizes the distributional hypothesis, i.e., `similar words tend to appear in similar contexts', by utilizing the context words to learn the representation of a given word. It is usually trained using a large corpus containing texts of similar domain. Since our dataset is much smaller, we used pretrained word2vec vectors, using a corpus containing 650K material science published papers, from the previous work of \citet{kim2017machine}.

Local context of the word occurring in the sentence proves useful in those cases where the word vector corresponding to a given token is not available. That is very common because of new materials and other tokens in the literature. Consider the following sentence in our dataset : \textit{We performed first principles calculations on \_\_\_ doped with Mn, focusing on different aspects.} From the context words, it is quite evident that the word that appears in place of blank belongs to the class of \textit{material}. To utilize the context information, we treat each sentence as a sequence of words that are represented using embedding. Since it is a sequence labelling task, we use recurrent neural networks (RNN) to learn the embedding of the context that the sentence represents \cite{mesnil2013investigation}. 

\begin{figure*}[hbt!]
    \centering
	\includegraphics[width=450pt]{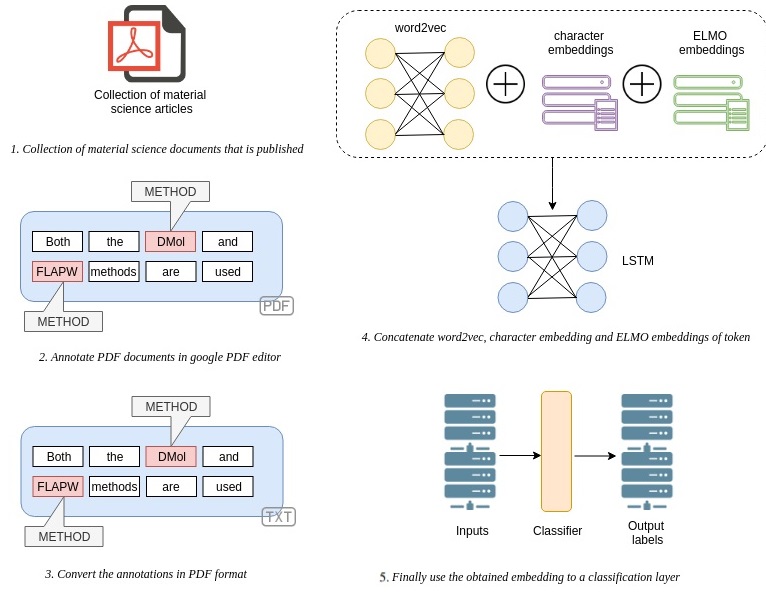}
	\vspace{-3mm}
	\caption{System workflow. Training process goes through the following stages : 1. Collect material science articles in PDF format from arxiv. 2. Annotate the articles using any PDF editor. 3. Convert the annotated PDF to text format that can be used for NLP tasks. 4. Train BiLSTM neural network model to learn the contextual embeddings of each token. 5. Pass the learned output embedding of LSTM to CRF layer to obtain the predicted label.}
	\label{fig:network}
\end{figure*}

RNN are known to suffer when the sequence is long as in that case it cannot utilize context words that are not very close to the target word. Because of this, we used Long Short Term Memory (LSTM) \cite{hochreiter1997long}, which is a variant of RNN. Forward LSTM can capture the sequence of the tokens appearing before the current token, while backward LSTM captures the sequence of the tokens appearing after the current token. While training we used Bidirectional LSTM (Bi-LSTM) \cite{huang2015bidirectional}, which is basically a concatenation of embedding obtained by forward LSTM and embedding obtained by backward LSTM to achieve this purpose. For each element in the input sequence, each LSTM layer computes the following
\begin{equation}
    i_{t} = \sigma({w_{i}[h_{t-1}, x_{t}] + b_{i}})
\end{equation}

\begin{equation}
    f_{t} = \sigma({w_{f}[h_{t-1}, x_{t}] + b_{f}})
\end{equation}

\begin{equation}
    o_{t} = \sigma({w_{o}[h_{t-1}, x_{t}] + b_{o}})
\end{equation}
where \(i_{t}\) represents the input gate, \(f_{t}\) represents forget gate, \(o_{t}\) represents output gate, \(\sigma\) represents sigmoid function, \(w_{t}\) represents weight of respective gate(x) neurons, \(h_{t-1}\) represents the output of previous block at time step \(t-1\), \(x_{t}\) represents input representation at current time-step, and \(b_{x}\) represents the biases for the respective gates. The equations to obtain the final output is

\begin{equation}
    \tilde{c_{t}} = tanh({w_{c}[h_{t-1}, x_{t}] + b_{c}})
\end{equation}

\begin{equation}
    c_{t} = f_{t} * c_{t-1} + i_{t} * \tilde{c_{t}}
\end{equation}

\begin{equation}
    h_{t} = o_{t} * tanh(c_{t})
\end{equation}
where \(c_{t}\) represents the cell state at time-step $t$ and \(\tilde{c_{t}}\) represents the candidate for cell state at time-step $t$. The output obtained at each time-step $t$ is passed through a fully connected linear layer to perform the task of classification among six classes. We applied softmax function to the logits obtained from the linear network to obtain the probabilities for each class.

Chemical formulae are often represented by an inorganic notation, for example - sodium sulphate is represented as Na2SO4. We can observe that this token has a fixed pattern where the first letter is uppercase, followed by lowercase characters and numbers. There are tokens with distinct shapes available in our dataset. Also, we notice that the prefix and suffix of a token play a major role in deciding the class to which it belongs. To detect the shape of the tokens, we treat each token as a sequence of characters and then apply Bi-LSTM. In the end, we obtain the character level embedding of the token.

After we obtain the embedding, we need to build a tag encoder that can rightly understand the class belonging to the token. For this purpose, we use a Conditional Random Field (CRF) classifier \cite{abramson2015sequence}.

 \citet{peters2018deep} showed that ELMO embedding of a token can give better performance as compared to Bi-LSTM embedding in general. In ELMO, each token is assigned a representation that is a function of the entire input sentence. ELMO embeddings are deep, in the sense that they capture the output of all the internal layers of LSTM. We used pretrained ELMO embedding for material science scientific articles. The input to the Bi-LSTM model is thus a concatenation of pretrained word2vec embedding, character embedding, and pretrained ELMO embedding (see Figure \ref{fig:model}).

The basic workflow of the model is shown in Figure \ref{fig:network}. 

We tried quite a few variations of the basic approach during our experiments, which we describe below.

\subsection{Varying the context length: Using sentence context vs. from an entire Section}
The length of the input sequence to the model largely determines the context available to correctly classify the tokens. As mentioned earlier, we aim to extract only the ``entities of interest" and not all the mentions from an article. In general, a sentence is an input to the model for performing NER. This is computationally favorable, but doing so results in loss of context, thus it is not suitable for our task. Instead of a sentence, we also experiment with the entire section as an input to the model. For example, if \textit{ABSTRACT} is a section, then all the text under this section becomes an input to the model. As we will see in the experimental results, using a larger context from the entire section yields a much better performance than restricting context to only at the sentence-level.  

\subsection{Relabeling and Mimicking Model}

In general, for training an NER model, only the sentences which contain at least one positive annotation, (i.e, a label other than $'$O$'$) are used. Discarding all other sentences from the training dataset, may result in loss of data, which could otherwise have been helpful. Also, this particular setting cannot be used for the purpose of section-wise training. To explore this, we perform experiments with two different settings. One with only sentences containing at least one annotation (dataset $d$), and other with all the sentences (dataset $D$). We then use our validation set to understand the performance of these settings. 

\begin{figure}[hbt!]
    \centering
	\includegraphics[width=\linewidth]{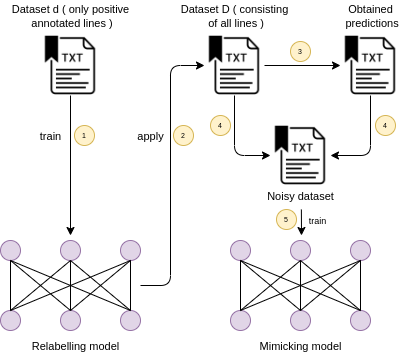}
	\caption{Relabeling and Mimicking model : 1. Relabeling model trained on dataset containing only positively annotated lines. 2. Applying the model on a dataset that contains all lines - both annotated and non-annotated. 3. Obtaining the predictions after applying the model. 4. Creating a noisy dataset that consists of the original labels and fraction of predicted labels from the previous step. 5. Train Mimicking model on the noisy dataset }
	\label{fig:RandM}
\end{figure}

Results show that model trained on dataset $d$, gives better result for \textit{material} and \textit{structure} class. Whereas, model trained on the dataset $D$ gives better result for \textit{method}, \textit{code} and \textit{parameter} class. These results can be explained by the frequency of these classes. As shown in the label distribution in figure \ref{fig:labelDistribution}, 
the number of spans of \textit{material} and \textit{structure} class is much less than that of other classes. The density of these labels further reduces with the addition of non-positive sentences in the dataset. It was observed that the first model had better recall scores and the second model had better precision scores. 

Our objective is to train a model that performs well for all the desired classes. For this we introduce, \textbf{Relabeling and Mimicking Model} (see Figure \ref{fig:RandM}). First, we train a model on the dataset $d$ containing only positively annotated lines. We call this model as the Relabeling model. Now, we add all the non-positive sentences in the dataset, thus the new dataset becomes $D$. Now, we use the Relabeling model to make prediction on this $D$ dataset only for \textit{material} and \textit{structure} class. Keeping all the positive ground truth labels for all the classes intact, we add some percentage of the new predicted positive labels in this dataset, as a way of adding meaningful noise. Thus, the new dataset now becomes $D'$. Now, we train a model on $D'$ dataset. We call this model as Mimicking Model. Increasing the amount of noise may increase the recall but decrease the precision. Different amounts of noise may be required for a different class, which has to be determined experimentally. As mentioned earlier a model trained on $D$ dataset has good precision but low recall. As we will see in the experiments, the mimicking model achieves good precision as well as recall scores.

\subsection{Post processing}
One very basic technique to improve the results of any NER task is to understand the pattern of occurrence of tokens across the classes by analyzing the dataset. We identified two such patterns for our dataset. i) the entities belonging to the class \textit{material} will generally also be present in the title of the paper and ii) entities belonging to class \textit{method} and \textit{parameter} are usually close to each other. To reduce the number of spurious entities obtained from the model, we apply the above-mentioned post-filters. We achieved this by removing the predicted materials that are obtained in any section other than TITLE. We also removed those \textit{methods} that does not contain any token classified as \textit{parameter} within a span of two sentences. Experimentally we observed that by applying these filters, the precision of these classes is increased without considerable decrease in recall. It is to be noted that for some articles, the PDF parsing application is not able to obtain the TITLE, thereby posing an issue to apply the first filter. We handled those cases by considering the ABSTRACT and initial two sections from the article, i.e., only those \textit{material} entities obtained in ABSTRACT and the initial two sections are considered and the rest are filtered out.

\subsection{Sentence Classification}
Recent state-of-the-art results in sentence classification, over multiple public datasets, are obtained by using the BERT model. It generates an embedding for each input instance and then we use it as an input to feed forward neural network to categorise it into \textit{positive} or \textit{negative}
classes. Training set used in this model is obtained by annotating the abstracts of 90 articles. We fine-tune the model using both \textit{bert-base-uncased} and \textit{scibert-base-uncased} pretrained embeddings. Experimental results show that training it using \textit{scibert-base-uncased} gave better F1 scores.

\section{Evaluation metrics}
\label{S:5}

We view the task of NER as classification of a token with one among the 5 classes that we are interested in. If some token does not belong to any such class, then we treat that token to belong to a special class denoted by ``O". As with any other classification tasks, we use precision, recall, and f-measure to measure the performance of our model. For each document we created two sets of tokens for each category - one set contains the actual annotations or ground truth (set \(GT\)) and the other set contains the predicted token (lets call it set \(Pred\)). Then we compared the tokens between the sets and used it to compute the metrics.

We define the precision (P) of a particular class (c) by the expression 
\begin{equation}
P = \frac{\mid GT \cap Pred \mid}{\mid Pred \mid}
\end{equation}
where \(GT\) represents the set that are actually labelled with \(c\) and \(Pred\) represents the set that is predicted as \(c\). Precision calculates the fraction of the obtained results that are relevant. We define recall (R) by the expression
\begin{equation}
R = \frac{\mid GT \cap Pred \mid}{\mid GT \mid}
\end{equation}
Recall calculates the fraction of total relevant results that re correctly classified by the system. In general, we want the system to output both high precision and high recall. In order to measure the accuracy of the system we use F1-score which is defined by the harmonic mean of precision and recall i.e 
\begin{equation}
F1 = \frac{2PR}{P + R}
\end{equation}

\begin{table*}[hbt!]
\centering
\renewcommand{\arraystretch}{1.1}
\begin{tabular}{|x{2cm}|x{1.95cm}|x{1.95cm}|x{1.95cm}|x{1.95cm}|x{1.95cm}|x{1.95cm}|}
\hline
 & \multicolumn{3}{x{7.5cm}|}{\textbf{Na\"{i}ve rule based method}} & \multicolumn{3}{x{7.5cm}|}{\textbf{Unsupervised Word embedding based method}} \\\hline
 & \textbf{P, $\sigma$} & \textbf{R, $\sigma$} & \textbf{F1-Score, $\sigma$} &  \textbf{P, $\sigma$} & \textbf{R, $\sigma$} & \textbf{F1-Score, $\sigma$} \\ \hline
\textit{\textit{material}} & 40.38, 16.3 & 20.35, 10.43 & 26.25, 11.58  &  64.17, 6.32 &	59.98, 11.04 &	61.72,7.84  \\ \hline
\textit{\textit{structure}} & 21.39, 9.84 & 29.79, 13.43 & 24.78, 11.38  & 38.62, 6.4	& 50.52,17.62 &	42.97,9.42   \\ \hline
\textit{\textit{method}} & 63.2, 3.62 & 69.03, 7.91 & 65.74,3.38 & 55.44, 4.72 & 85.95,1.52 &	67.27,3.11  \\ \hline
\textit{\textit{parameter}} & 7.78, 0.34 &	98.36, 1.32 & 14.43, 0.59 & 18.67, 7.01 & 88.3,6.27 &	30.16,9.13  \\ \hline
\textit{\textit{code}} & 84.74, 2.66 & 60.6, 7.62 & 70.42, 5.08 &  84.75, 7.12 &	72.97, 12.76 & 77.43,5.46  \\ \hline
\textbf{macro score} &  43.3, 4.1 & 55.63, 5.85 & 41.76, 7.55  & 52.33, 4.6 & 71.54, 5.9 &	55.84, 4.89  \\ \hline
\end{tabular}
    \caption{Baselines - Precision (P), Recall (R), F1-Score and their respective standard deviations ($\sigma$) for Na\"{i}ve rule based method and Unsupervised word embedding based method (UWE)~\cite{tshitoyan2019unsupervised} }
  \label{tab:basemodel1}
\end{table*}

\begin{table*}[b]
\centering
\renewcommand{\arraystretch}{1.1}
\begin{tabular}{|x{2cm}|x{1.95cm}|x{1.95cm}|x{1.95cm}|x{1.95cm}|x{1.95cm}|x{1.95cm}|}
\hline
 & \multicolumn{3}{x{7.5cm}|}{\textbf{BERT question answering model}} & \multicolumn{3}{x{7.5cm}|}{\textbf{BERT domain adaptation model}} \\\hline
 & \textbf{P, $\sigma$} & \textbf{R, $\sigma$} & \textbf{F1-Score, $\sigma$} &  \textbf{P, $\sigma$} & \textbf{R, $\sigma$} & \textbf{F1-Score, $\sigma$} \\ \hline
\textit{\textit{material}} & 71.82, 3.51 & 73.56, 4.23 & 72.68, 5.72  &  76.16, 4.03 & 58.48, 5.02 & 66.29, 4.02  \\ \hline
\textit{\textit{structure}} &  46.51, 3.93 & 35.21, 5.34 & 40.08, 4.31  &   45.58, 7.14 & 40.29, 9.81 & 42.29,7.33  \\ \hline
\textit{\textit{method}} & 60.62, 2.56 & 71.23, 3.45 & 65.50, 3.76 & 81.11, 2.27 &	63.97, 3.55 & 71.47,2.26  \\ \hline
\textit{\textit{parameter}} & 62.51, 2.87 &	65.72, 6.31 & 64.07, 5.38 &  81.04, 3.81 & 73.02, 5.22 & 76.74, 3.7  \\ \hline
\textit{\textit{code}} & 82.11, 3.13 & 71.89, 2.35 & 76.66, 3.29 & 82.74, 4.34 & 80.63, 3.7 & 81.55, 2.02  \\ \hline
\textbf{macro score} & 64.71, 4.52 & 63.52, 5.68 & 63.79, 5.14   & 73.33, 2.3 & 63.28, 2.54 & 67.65, 2.48 \\ \hline
\end{tabular}
    \caption{Baselines - Precision (P), Recall (R), F1-Score and their respective standard deviations ($\sigma$) for BERT question answering model and BERT domain adaptation model}
  \label{tab:basemodel2}
\end{table*}

A model with a high F1-score is considered better as compared to a model with lower F1-score. In our task, we are concerned with \textit{spans} of tokens. In order to calculate how well the model is performing, we have followed combination of two different approaches: a) exact match and b) partial match \cite{breck2007identifying}. Revised expression to calculate precision and recall is given by
\begin{equation}
precision = \frac{\mid \{p : p \, \exists \, Pred \wedge a \, \exists \, GT \, \wedge \, \, \alpha (a,p)\} \mid}{\mid Pred \mid}
\end{equation}
 
\begin{equation}
recall = \frac{\mid \{a : p \, \exists \, Pred \wedge a \, \exists \, GT \, \wedge \, \, \alpha (a,p)\} \mid}{\mid GT \mid}
\end{equation}

For exact matching, \( \alpha(a,p) \) is true if the \(a\) and \(p\) represents exactly the same spans, whereas in partial matching, \( \alpha(a,p) \) is true if there is some overlap between the spans of \(a\) and \(p\).

\begin{equation}
\sigma=\sqrt{\frac{1}{N-1}\sum_{i=1}^{N} (X_{i} -\bar{X})^2} 
\label{eq:sd}
\end{equation}

In our experiments, we calculated the average precision (P), recall (R) and F1-score along with their respective standard deviations ($\sigma$) where $\sigma$ values were calculated as per Equation \ref{eq:sd}. We have utilized string match as an accuracy metric. In other words, for each article, we compared the list of annotated tokens for each class with the list of predicted tokens for that class. This is significantly different than that of usual NER setting where we are interested in finding the class level metrics for each sentence.

\section{Experimental Results}
\label{S:6}

The task of extracting the entities from material science articles shares some similarities with the Named Entity Recognition task (NER) and also with question-answering task \cite{zhang2017exploring}.  
Thus, we consider state-of-art models for these tasks as the baseline. For NER, we followed the work of \citet{beltagy2019scibert} and used Scibert which is pretrained on scientific publications. Since our task is specific to the material science domain, we used unsupervised domain adaptation techniques to obtain a BERT model specific to the material science domain as discussed by \citet{han2019unsupervised}. For the question-answering task, we used the BERT multi-answer question answering model showed by \citet{devlin2018bert}. Along with the above two models, we also used a baseline that predicts based on the frequency of occurrence in the training set. Thus, in the test dataset, if a span is present in the training dataset with a class annotation, it is given that particular label. We call this as the Na\"{i}ve rule-based approach. Another baseline, we have taken is the unsupervised word embedding based method (Word2Vec), developed by Tshitoyan et al.~\cite{tshitoyan2019unsupervised}. The results for Na\"{i}ve rule-based approach and Unsupervised Word embedding based method (UWE)~\cite{tshitoyan2019unsupervised} are listed in Table \ref{tab:basemodel1}. The results for BERT question answering model and BERT domain adaptation model are listed in Table \ref{tab:basemodel2}. We have performed 5 fold cross validation test for each model to report average precision (P), recall (R) and F1-score (F1) along with their standard deviations ($\sigma$) across five folds.

\begin{table*}[hbt!]
\centering
\renewcommand{\arraystretch}{1.1}
\begin{tabular}{|x{2cm}|x{1.95cm}|x{1.95cm}|x{1.95cm}|x{1.95cm}|x{1.95cm}|x{1.95cm}|}
\hline
 & \multicolumn{3}{x{7.5cm}|}{\textbf{Per-line setup with only positively annotated lines model}} & \multicolumn{3}{x{7.5cm}|}{\textbf{All lines per section model}} \\\hline
 & \textbf{P, $\sigma$} & \textbf{R, $\sigma$} & \textbf{F1-Score, $\sigma$} &  \textbf{P, $\sigma$} & \textbf{R, $\sigma$} & \textbf{F1-Score, $\sigma$} \\ \hline
\textit{\textit{material}} & 78.31, 8.92 & 85.4,5.47 & 81.38,5.28 & 80.54, 6.1 & 78.43, 12.81 & 78.92, 7.38 \\ \hline
\textit{\textit{structure}} & 35.45, 4.23 &	61.02, 20.03 & 43.58, 7.19  & 45.16, 16.98 & 63.5, 15.65 & 52.78, 10.3  \\ \hline
\textit{\textit{method}} & 59.68, 9.54 &	89.34, 4.03 & 71.07, 6.77 &  65.5, 7.91 &	81.91, 4.21 & 72.46, 4.9  \\ \hline
\textit{\textit{parameter}} & 52.39, 6.0 & 85.05, 1.68 & 64.64, 4.31  & 62.85, 11.27 & 81.34, 6.94 & 70.11, 6.25  \\ \hline
\textit{\textit{code}} & 69.16, 10.55 &	85.74, 5.05 & 75.98, 5.27  & 	 73.97, 14.69 & 82.88, 7.47 & 77.16, 6.21 \\ \hline
\textbf{macro score} & 58.99, 4.02 & 81.31, 6.0 & 67.43, 1.52  & 65.61, 9.34 & 77.61, 7.48 & 70.29, 3.97  \\ \hline
\end{tabular}
    \caption{Precision (P), Recall (R), F1-Score and their respective standard deviations ($\sigma$) for Bi-LSTM-CRF on different context length and different sections}
  \label{tab:linevssection}
\end{table*}

\begin{table*}[b]
\centering
\renewcommand{\arraystretch}{1.1}
\begin{tabular}{|x{2cm}|x{1.95cm}|x{1.95cm}|x{1.95cm}|x{1.95cm}|x{1.95cm}|x{1.95cm}|}
\hline
 & \multicolumn{3}{x{7.5cm}|}{\textbf{Per-section setup with only positively annotated lines (Relabeling Model)}} & \multicolumn{3}{x{7.5cm}|}{\textbf{Mimicking model}} \\\hline
 & \textbf{P, $\sigma$} & \textbf{R, $\sigma$} & \textbf{F1-Score, $\sigma$} &  \textbf{P, $\sigma$} & \textbf{R, $\sigma$} & \textbf{F1-Score, $\sigma$} \\ \hline
\textit{\textit{material}} & 80.23, 3.91 & 89.03, 6.29 & 84.05, 3.09 & 80.7, 5.72 & 87.27, 3.71 & 83.66, 1.9   \\ \hline
\textit{\textit{structure}} & 38.39, 8.6 & 75.32, 14.04 &	50.81, 10.53   & 46.14, 16.11 &	67.67, 12.22 & 53.75,11.28    \\ \hline
\textit{\textit{method}} & 49.6, 5.21 & 	92.57, 2.28 & 64.5, 4.82 & 66.24, 5.62 & 88.01, 5.02 & 75.44, 3.93  \\ \hline
\textit{\textit{parameter}} & 39.49, 7.98 & 88.31, 2.39 &	54.1,7.65   & 69.51, 9.53 & 82.26, 2.76 & 74.95, 4.53  \\ \hline
\textit{\textit{code}} &  68.31, 5.77 &	90.04, 0.86 & 77.57, 3.51  & 74.19, 9.17 &	84.05, 4.23 & 78.4, 3.46  \\ \hline
\textbf{macro score} & 55.2, 3.51 &	87.06, 3.05 & 66.55, 3.48  & 67.35, 8.6 & 81.85, 2.92 & \textbf{73.13}, 4.66  \\ \hline
\end{tabular}
    \caption{Precision (P), Recall (R), F1-Score and their respective standard deviations ($\sigma$) for Relabeling and Mimicking model}
  \label{tab:final}
\end{table*}

Next, we compare the results of the per-line training (with only positively annotated lines) and all lines per-section training setup. In Section 4.2 we mentioned that we used the section-wise training setup. In Table \ref{tab:linevssection}, we compare the results of these two training setups. The model trained using the section-wise training setup outperforms the model trained on a sentence-wise training setup. 

In Section 4.3 we mentioned that using only positive-annotated lines for training can lead to information loss. Including all lines per section (positive and negative) in the training, dataset leads to comparatively better scores as seen in Table \ref{tab:linevssection}. Since we have already established that section-wise training is better, we adopted this approach in both the models. Only the training dataset for them is different.  
Results in Table \ref{tab:linevssection} show that the model trained on only positive-annotated lines do have good recall scores, but low precision scores and the model trained on all lines have good precision scores but low recall scores. Also, except \textit{material} for all other classes, the later model performs better than the former in terms of F1 score.

We have already explained the Relabeling and Mimicking model in detail in section 4.3. The model trained on only positive-annotated lines (section-wise) is used as the Relabeling model due to its high recall scores. Relabeling model is similar to the per-section setup with only positively annotated lines. In Table \ref{tab:final} we compare the results of the Relabeling model and the Mimicking model. The mimicking model achieves a better macro F1-score as well as has the best F1 for all the classes. We have tuned several hyper-parameters of the model to get the optimum results. In our case, we have set ``$learning\_rate$" = ``0.001", ``optimizer" = ``sgd", ``dropout" = ``0.5", ``$hidden\_dim$" = ``200", ``$num\_epochs$" = ``120", ``$batch\_size$" = ``8" to get the best F1-score.

We use the Relabeling model to add noise in the training dataset of the Mimicking model. Keeping the ground truth labels intact, we add some percentage of the extra labels predicted by the Relabeling model. The amount of noise added per class is shown in Table \ref{tab:noise}. For example, if 50 \% noise is added to the \textit{material} class, it means half of the extra predicted labels of this class are added into the training data randomly. The amount has to be balanced because increasing noise results in an increase in the recall but a decrease in precision. The optimum noise level has to be determined experimentally. We used the validation set results to fine-tune the noise level for each individual classes (as shown in Table \ref{tab:noise}).

\begin{table}[ht]
\centering
\begin{tabular}{|c|c|}
\hline
\textbf{Class} & \textbf{\% Noise} \\ \hline
\textit{material}  & 50       \\ \hline
\textit{structure} & 50       \\ \hline
\textit{method}    & 25       \\ \hline
\textit{parameter} & 0        \\ \hline
\textit{code}      & 33       \\ \hline
\end{tabular}
\caption{Noise percentage in the Mimicking model}
\label{tab:noise}
\end{table}

\begin{figure}[!hbt]
	\centering
	\includegraphics[width=\linewidth]{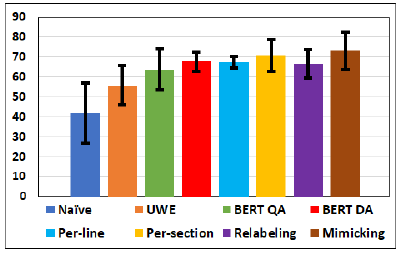}
	\caption{Performance of the model (F1-Score) with confidence interval of 95\%}
	\label{fig:f1-score}
\end{figure}

We have shown summary of F1-scores of all models along with their 95\% confidence interval (error bars are shown in black) in Figure \ref{fig:f1-score}. It is seen that the Mimicking model achieves the best F1-score performance among all approaches - Na\"{i}ve rule based (NB), Unsupervised Word Embedding (UWB), BERT Question Answering (BERT QA), BERT domain adaptation (BERT DA), Per-line, Per-section (All lines), Relabeling and Mimicking model.

We observe that all the competing models fail to identify micro details of the structures properly. Thus, the precision, recall and F1-scores for structure are lower than other classes. This is due to the fact that systems correctly identifies macro details of structures like ``tetragonal", ``cubic" etc. but not able to match micro details of the structures like - ``cubic perovskite Pm30304m".

\begin{table}[tbh!]
\begin{tabular}{|l|l|l|}
\hline
Class & Entity & F1-Score \\\hline
\multirow{5}{*}{Methods}  & GGA (Generalised   & 96.67\\ 
& Gradient Approximation) & \\\cline{2-3} 
 & DFT (Density  & 94.56 \\ 
 & Functional Theory) &\\\cline{2-3} 
                  & PAW (Projected  & 93.31\\
                  &  Augmented Wave) & \\ \cline{2-3} 
                  & PBE (Perdev  & 94.54 \\
                  &  Burke and Ernzerhof) &\\ \cline{2-3} 
                  & FP-LAPW (full potential   & 91.10 \\
                  & linearized augmented plane wave &\\ \hline
\multirow{5}{*}{Parameters} & eV  & 97.56\\ \cline{2-3}
                  & K-point  & 84.06 \\ \cline{2-3} 
                  & Basis Set  & 76.16 \\ \cline{2-3} 
                  & gmax &  87.43 \\ \cline{2-3} 
                  & lmax &  89.81 \\ \hline

\end{tabular}
\caption{F1-Scores of top five frequent extracted methods and parameters}
\label{tab:top5methAndparam1}
\end{table}

\subsection*{Detailed Analysis}
To perform a detailed analysis of the results, we take our best method, ``mimicking model'' and analyze the individual accuracy scores while identifying the top five most frequent methods and parameters. The F1-score for each of these entities is shown in Table \ref{tab:top5methAndparam1}. We see that our approach is able to extract several entities (methods like GGA, DFT, PAW, PBE, LAPW and parameters like eV, K-points, Basis Set, gmax, lmax)  with very good F1-score.

\subsection*{Summary Extraction}
For summary extraction, results obtained after applying the BERT model are shown in Table \ref{tab:bertbase}. We used the generic BERT ({\sl bert-base-uncased}) as well as SCI-BERT ({\sl scibert-scivocab-uncased}), and found that the generic BERT gives much better results in terms of F1-score.

\begin{table}[hbt!]
\centering
\renewcommand{\arraystretch}{1.2}
\begin{tabular}{|x{3.5cm}|x{1.2cm}|x{1.2cm}|x{1.2cm}|}
\hline
\textbf{Model} & \textbf{P} & \textbf{R} & \textbf{F1} \\ \hline
BERT with \textit{scibert-scivocab-uncased} pretrained model & 90.91 & 71.43 & 80.00 \\ \hline
BERT with \textit{bert-base-uncased} pretrained model & 82.61 & 95.00 & \textbf{88.37} \\ \hline
\end{tabular}
    \caption{Macro-averaged scores obtained from sentence classification model using pretrained model}
  \label{tab:bertbase}
\end{table}

\section{Sample Output}
\label{S:7}
In this section we present some of the actual outputs obtained from the best model for the information extraction task.

Tables \ref{tab:output1} and \ref{tab:output2} show the output obtained by applying the best model on material science papers titled, \textit{``Single pair of Weyl fermions in thehalf-metallic semimetal EuCd2As2"} and \textit{``Electron-phonon superconductivity in CaBi2 and the role of spin-orbit interaction"}, respectively. 

\begin{table}[hbt!]
\centering
\renewcommand{\arraystretch}{1.2}
\begin{tabular}{|x{1.2cm}|x{2.8cm}|x{2.8cm}|}
\hline
\textbf{Class} & \textbf{Ground truth} & \textbf{Model output} \\ \hline
\textit{material}  
&
• EuCd2As2 
&
• half - metallic semimetal EuCd2As2 \\ \hline
\textit{structure} 
& 
• space group 164(P3mm) 
& 
• space group 164(P3mm)\newline • hexagonal \\ \hline
\textit{method}    
& 
• PBE\newline • Generalized Gradient Approximation 
& 
• Perdew - Burke - Ernzerhof (PBE)\newline
• projector - augmented wave [37] method\newline
• Generalized Gradient Approximation \\ \hline
\textit{parameter} 
& 
• 318 eV\newline • U = 5.0 eV\newline • 0.01 eV/Å 
& 
• 318 eV\newline • U = 5.0 eV\newline • 0.01 eV/Å \\ \hline
\textit{code} 
& 
• VASP 
& 
• VASP\\ \hline
\end{tabular}
\caption{Outputs obtained on paper titled \textit{``Single pair of Weyl fermions in the half-metallic semimetal EuCd2As2"}}
\label{tab:output1}
\end{table}

\begin{table}[hbt!]
\centering
\renewcommand{\arraystretch}{1.2}
\begin{tabular}{|x{1.2cm}|x{2.8cm}|x{2.8cm}|}
\hline
\textbf{Class} & \textbf{Ground truth} & \textbf{Model output} \\ \hline
\textit{material}  
& 
• CaBi2 
& 
• CaBi2 \\ \hline
\textit{structure} 
& 
• space group Cmcm, No. 63\newline 
• cubic fcc 
&
• space group Cmcm, No. 63\newline 
• cubic \\ \hline
\textit{method}    
& 
• densityfunctional theory(DFT)\newline
• Perdew - Burke - Ernzerhof\newline 
• Rappe-Rabe-Kaxiras-Joannopoulos ultrasoft pseudopotentials 
&
• densityfunctional theory(DFT)\newline
• pseudopotential method\newline
• Perdew - Burke - Ernzerhof\newline
• Rappe-Rabe-Kaxiras-Joannopoulos ultrasoft pseudopotentials\\ \hline
\textit{parameter} 
&
• grid of 123 k points\newline 
• grid of 43 q points
&
• 123 k points\\ \hline
\textit{code} 
& 
• QUANTUM ESPRESSO\newline
• BOLTZTRAP
& 
• QUANTUM ESPRESSO\newline
• BOLTZTRAP\\ \hline
\end{tabular}
\caption{Outputs obtained on paper titled \textit{``Electron-phonon superconductivity in CaBi2 and the role of spin-orbit interaction"}}
\label{tab:output2}
\end{table}

\section{Post-facto analysis}
\label{S:8}
\begin{figure*}[!hbt]
	\centering
	\includegraphics[width=\linewidth]{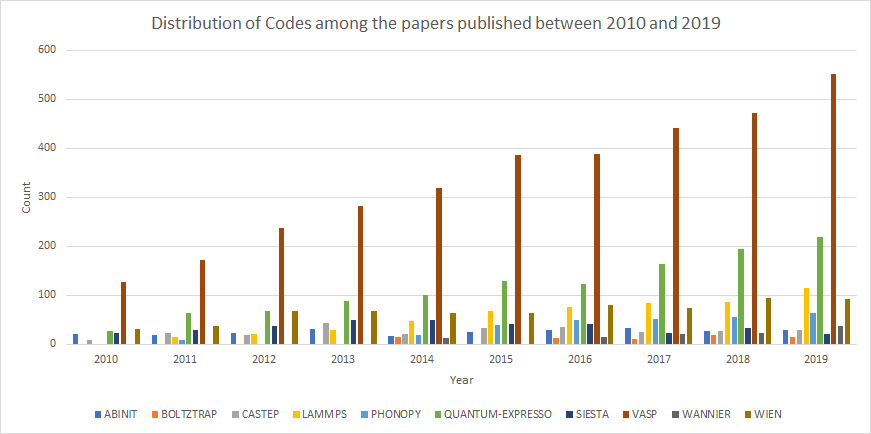}
	\caption{Distribution of codes among papers published between 2010 to 2019}
	\label{fig:code}
\end{figure*}

\begin{figure*}[!hbt]
	\centering
	\includegraphics[width=\linewidth]{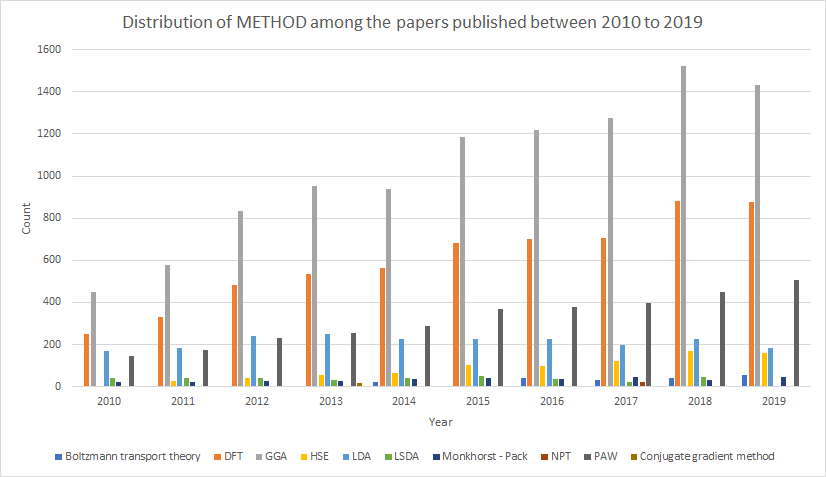}
	\caption{Distribution of methods among papers published between 2010 to 2019}
	\label{fig:method}
\end{figure*}

The proposed model can be used to observe various statistics over time. For instance, what are the main methods and codes being used over time? To answer this question, we run our best model on the 10K material science articles from arxiv (see Section \ref{S:3.1}). Figures \ref{fig:code} and \ref{fig:method} show the distribution of \textit{code} and \textit{method}, respectively over published articles in the last ten years. We can see from the Figure that ``VASP" is the most popular code of the decade followed by ``QUANTUM ESPRESSO''. Both codes have a common trend, their use has grown over the years, whereas the use of post-processing packages, such as ``Phonopy'' and ``Boltztrap'', has fluctuated over time. For the method tag, we see that ``GGA" appears the most followed by ``DFT" and ``PAW". This is indeed true that most of the researchers in the field of computational materials science use VASP as their primary choice as code and the most popular approach is PAW-PBE. We recognized that ``eV" is the most popular parameter followed by ``K-point". Therefore, we also would like to point out that this utility would be useful to track the choices of the researchers in this field over the time globally.

\section{Online interface}
\label{S:9}
To be able to make the proposed methods accessible and usable by the community, we have created the online interface: 

server: \url{http://34.69.41.173:3333/}. 

\noindent With the help of this website, one can upload any published paper in PDF format and obtain the predictions using our best model. It supports batch upload of multiple PDF documents at the same time. This website will allow the material science researchers to automatically extract entities of interest from uploaded papers. We expect this will enhance the pace of future material science researches. The server side of the application is developed using python Django~\footnote{\url{https://djangoproject.com}} and the model is trained using PyTorch package~\cite{paszke2019pytorch}. 

\section{Conclusions}
\label{S:10}
This paper explored the task of extracting relevant entities from material science publications. We annotated 214 material science articles with five classes - \textit{material, code, method, structure and parameter} and used the annotated data to train a deep neural network (see Section \ref{S:4}). We applied our trained model to obtain entities from a set of 10K material science articles that are published in arxiv over the last ten years. Additionally, we trained a sentence classification model to extract sentences appearing under \textit{abstract} section that are related to the results. This information helped us to generate a short summary level information of the published article. By using our model, we pulled out statistics as shown in Section \ref{S:8} giving an insight of the usage of \textit{methods and codes} in a span of ten years. We created an online interface that lets us obtain the predicted entities from an uploaded published article (see Section \ref{S:9}). Using our tool, one can index the material science documents according to the material used or parameter used for the specific methods. 

Most of the sentences in an article will not contain any entity that we are interested in. Passing these sentences to the system is of no use since the model will not output any entity from those. One possibility is to automatically identify relevant sections of the text data and pass as input only the sentences that are contained in those sections. In order to achieve this, we need to first classify each section of the document as whether it is relevant or not. As a future work, we can build a classifier that can classify sentences or sections of text as relevant or not, and then feed this input to the model for prediction.  
Currently, we are only extracting text information from the PDF documents. A further improvement can be obtained if we extract figures and tables from published documents. Material science documents contain useful information about
materials and the output obtained from those materials using different methods in
the form of tables. We expect most entities to appear in the table section. Hence extracting those information will enable us to collect more information. An additional feature can be to correctly identify the name of the parameter for every span that is predicted as the class parameter. From the dataset, it is clear that the parameter name occurs in the same sentence of the predicted token. We may use POS tags or dependency trees to accurately output the parameter name as well.

\section{Data Availability}
The code and data are available in \url{https://github.com/TeamMatSciE/MatSciE}.






\bibliographystyle{elsarticle-num-names}
\bibliography{sample-v1.bib}







\end{document}